\documentclass{article}

\usepackage[preprint]{corl_2026} 

\usepackage{amsmath,amssymb}
\usepackage{graphicx}
\usepackage{booktabs}
\usepackage{multirow}
\usepackage{enumitem}
\usepackage{hyperref}
\usepackage{caption}
\usepackage{subcaption}
\usepackage{float}
\usepackage{placeins}
\usepackage{pifont}

\usepackage{makecell}
\usepackage[table]{xcolor}
\usepackage{adjustbox}

\usepackage{graphicx}
\usepackage{xcolor}
\usepackage{pgfplots}
\usepackage{wrapfig}

\pgfplotsset{compat=1.18}
\definecolor{rankone}{RGB}{252,187,161}   
\definecolor{ranktwo}{RGB}{254,217,118}   
\definecolor{rankthree}{RGB}{255,247,188} 
\newcommand{\first}[1]{\cellcolor{rankone}\textbf{#1}}
\newcommand{\second}[1]{\cellcolor{ranktwo}#1}
\newcommand{\third}[1]{\cellcolor{rankthree}#1}

\title{\textbf{SplitAdapter: Load-Aware Humanoid Loco-Manipulation via Factorized Adaptation}}

\author{Jeonguk Kang \quad Hanbyel Cho \quad Sanghyun Kang \quad Donghan Koo \\
Future Robot AI Group, Samsung Electronics \\
Page: \href{https://splitadapter.github.io/}{https://splitadapter.github.io/}
}

\begin{document}
\maketitle


\begin{figure}[H]
    \centering
    \includegraphics[width=1.0\columnwidth]{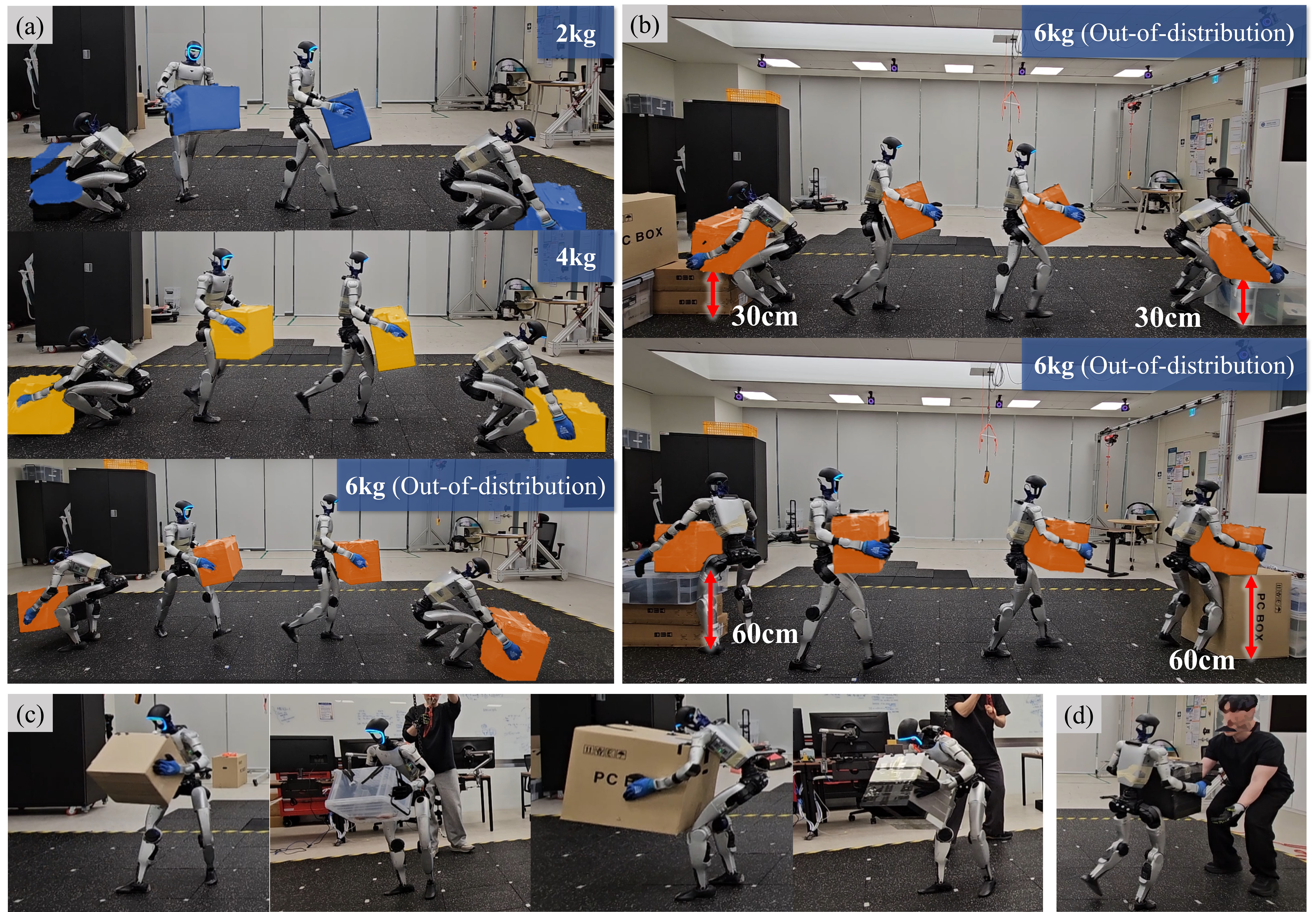}
    \caption{Real-world humanoid loco-manipulation experiments with the proposed method under diverse payload, pickup/placement-height, and object-interaction conditions. (a) Transport under different payload masses, including a 6 kg condition outside the training mass range. (b) Heavy-load (6 kg) transport under different pickup/placement heights (30 cm and 60 cm). (c) Generalization to varied object properties, including large boxes and acrylic boxes. (d) Human-object handover.}
    \label{fig:firstpage}
\end{figure}

\begin{abstract}
Humanoid loco-manipulation requires stable whole-body control under varying object masses and pickup/placement heights. This becomes particularly challenging in sim-to-real transfer, where object-induced load variation and robot-side dynamics mismatch interact during physical contact. Existing history-based adapters often compress these factors into a single latent representation, which can weaken robustness under heavy-load manipulation. We propose \textbf{SplitAdapter: Load-Aware Humanoid Loco-Manipulation via Factorized Adaptation}, which freezes a pretrained box manipulation policy and extends it with object/load and dynamics-aware context encoders trained with split world-model objectives, GRL-based cross-adversarial regularization, and hierarchical Feature-wise Linear Modulation (FiLM). In sim-to-sim experiments and real-world deployment, SplitAdapter improves Full-task success over the base policy and world-model FiLM baselines across object masses of $2$, $4$, and $6$ kg and pickup/placement heights of $0$, $30$, and $60$ cm, with the largest improvements under heavy-load conditions.
\end{abstract}

\keywords{Humanoid Loco-Manipulation, Sim-to-Real, Online Adaptation}

\section{Introduction}

Recent learning-based systems have demonstrated strong humanoid loco-manipulation capabilities~\cite{deepmimic,amp,phys_hsi,ultra,prohoi,resmimic,falcon,haic}, but sim-to-real transfer remains difficult under payload variation and robot/environment dynamics mismatch.
A heavier box or lower pickup height changes the required whole-body coordination, while actuator or dynamics mismatch can cause policies to behave differently on real hardware.

Existing adaptation methods often encode recent observation history into a single latent representation through supervised extrinsics estimation~\cite{rma}, privileged-state reconstruction~\cite{dwl}, or world-model prediction~\cite{wmr,rwm,any2track}.
While effective for locomotion and disturbance adaptation, these approaches can mix load-related information with residual dynamics mismatch, limiting robustness under heavy-load transfer.

We propose \textbf{SplitAdapter: Load-Aware Humanoid Loco-Manipulation via Factorized Adaptation}, a factorized adaptation framework for humanoid box lifting, transport, and placement.
Instead of retraining the full controller, SplitAdapter freezes a pretrained AMP~\cite{amp}-style manipulation policy and learns factorized adaptation modules on top of it.
The framework uses two context branches: an object/load branch that estimates object mass and loaded state while extracting object-related context, and a dynamics-aware branch that learns a dynamics-aware latent through robot-transition prediction.
The two branches are trained with split world-model objectives and GRL~\cite{grl}-based cross-adversarial regularization to reduce overlap between load-related and dynamics-related signals.
The resulting context representations modulate the frozen policy through hierarchical Feature-wise Linear Modulation (FiLM~\cite{perez2018film}).

We validate SplitAdapter in MuJoCo~\cite{mujoco} sim-to-sim experiments and real-world humanoid deployment under diverse object masses and pickup/placement heights.
Compared with the base policy and world-model FiLM baselines, SplitAdapter improves Lift-up success and Full-task success, particularly under heavy-load conditions.

\noindent In summary, our main contributions are:
\begin{itemize}[leftmargin=1.2em, topsep=2pt, itemsep=1pt, parsep=0pt, partopsep=0pt]
    \item We introduce SplitAdapter, a frozen-policy adaptation framework for humanoid box loco-manipulation under payload and dynamics variations.
    \item We propose a factorized adaptation structure that decomposes adaptation into object/load and dynamics-aware context using split world-model and GRL-based cross-adversarial separation.
    \item We validate SplitAdapter in MuJoCo sim-to-sim transfer and real-world humanoid deployment, demonstrating improved heavy-load loco-manipulation across diverse object masses, pickup/placement heights, and interaction scenarios.
\end{itemize}

\section{Related Work}

\paragraph{Humanoid loco-manipulation and object interaction.}
Early model-based approaches studied humanoid walking, multi-contact motion planning, and whole-body physical interaction using preview control, trajectory optimization, and hierarchical control~\cite{kajita_zmp,murooka_locomani,ruscelli_multicontact}.
More recent work has explored humanoid box loco-manipulation and heavy-object interaction through sim-to-real learning and optimization-based whole-body control~\cite{dao_box_locomani,rigo_heavy_objects}.
Learning-based humanoid-object interaction methods have further improved generalization and real-world performance.
Wang et al. proposed PhysHSI for real-world humanoid-scene interaction with AMP-based policy learning and coarse-to-fine perception~\cite{phys_hsi};
He et al. introduced ULTRA, a unified multimodal controller for autonomous humanoid whole-body loco-manipulation~\cite{ultra};
Lin et al. proposed Pro-HOI with root-guided commands and persistent object estimation for long-horizon box carrying~\cite{prohoi};
Zhao et al. introduced ResMimic, which adapts a pretrained general motion tracker with residual policies for object-aware loco-manipulation~\cite{resmimic};
and Zhang et al. proposed FALCON for force-adaptive humanoid loco-manipulation through upper- and lower-body decomposition~\cite{falcon}.
Other recent systems study video-based interaction learning and demonstration-based generalization~\cite{hdmi,omni_retarget,visualmimic,intermimic,demo_hlm}.
In contrast, SplitAdapter focuses on load-aware adaptation for heavy humanoid box manipulation through a frozen-policy adapter structure.

\paragraph{Sim-to-real transfer and world-model-based adaptation.}
Online adaptation for legged robots often estimates latent environment parameters from recent history.
Kumar et al. proposed RMA for rapid motor adaptation using estimated extrinsics~\cite{rma}, and later extended this idea to bipedal robots through A-RMA~\cite{arma_biped}.
Li et al. further showed that history-based reinforcement learning can produce versatile and robust bipedal locomotion controllers that adapt to dynamics changes through input-output history~\cite{biped_rl}.
Dynamics randomization and asymmetric actor--critic training are also widely used tools for deployable policies~\cite{peng_sim2real,pinto_asym}.
World-model-based methods use prediction or reconstruction as auxiliary objectives for learning adaptation embeddings.
Gu et al. proposed DWL to learn robust humanoid locomotion with denoising world-model learning~\cite{dwl}, and Sun et al. proposed WMR to reconstruct privileged states for humanoid locomotion~\cite{wmr}.
Li et al. studied neural dynamics models for robot policy optimization in RWM~\cite{rwm}.
Zhang et al. used world-model prediction to learn adaptation embeddings for motion tracking under disturbances in Any2Track/AnyAdapter~\cite{any2track}.
SplitAdapter follows this predictive-adaptation direction but separates object/load and robot-transition targets instead of compressing all adaptation factors into a single latent.
We also use GRL-based adversarial regularization~\cite{grl} to reduce information leakage between the two adaptation latents.

\section{Methodology}
\label{sec:methodology}


SplitAdapter is a load-aware factorized adaptation framework built on top of a frozen PhysHSI-style AMP-based humanoid box manipulation policy~\cite{phys_hsi}. It estimates object/load and dynamics-aware contexts from recent interaction history and injects them into the frozen policy through hierarchical Feature-wise Linear Modulation (FiLM)~\cite{perez2018film}.

\subsection{Frozen Base Policy}
We train the base policy with an AMP-based reinforcement learning framework for humanoid box manipulation, where target joint positions are executed by a low-level PD controller.
After pretraining, the policy is frozen and only the adaptation modules are optimized.

\begin{figure}[t]
    \centering
    \includegraphics[width=\textwidth]{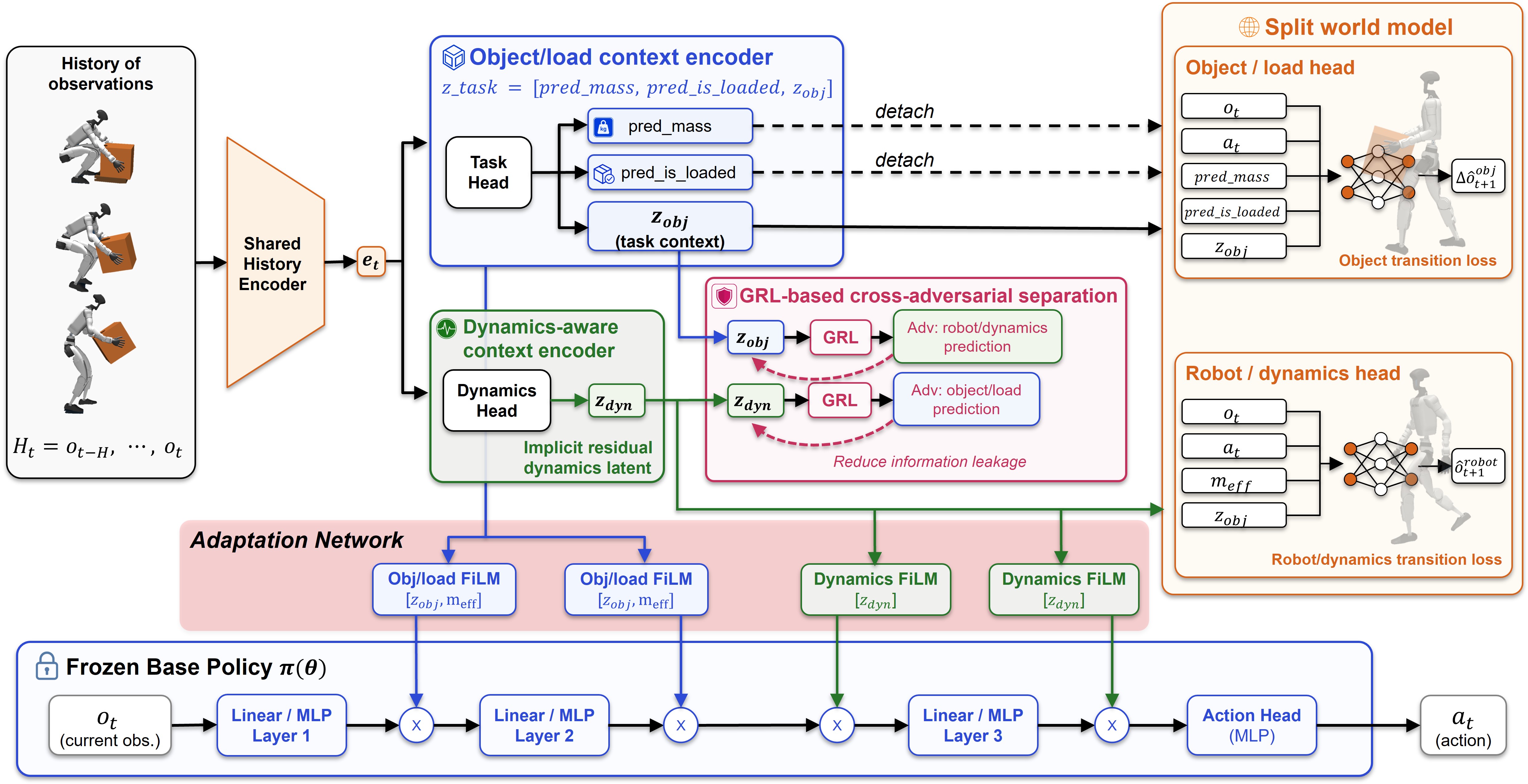}
    \caption{
    SplitAdapter overview. A frozen humanoid manipulation policy is adapted with object/load and dynamics-aware context encoders. The resulting factorized latents are trained with split world-model objectives, regularized with GRL-based separation, and injected through hierarchical FiLM modulation.
    }
    \label{fig:ocean_overview}
\end{figure}

\subsection{Dual-Context History Encoding}
To avoid compressing object/load and robot-side dynamics effects into a single latent, SplitAdapter factorizes adaptation into object/load and dynamics-aware context branches.

At time $t$, the adapter observes a recent observation-action history
\begin{equation}
    \mathcal{H}_t =
    \{o_{t-H}, a_{t-H}, \ldots, o_{t-1}, a_{t-1}, o_t\},
\end{equation}
where $H$ is the history length, and $o_t$ and $a_t$ denote the policy observation and action.
A shared history encoder first extracts a compact representation:
\begin{equation}
    e_t = f_{\mathrm{hist}}(\mathcal{H}_t).
\end{equation}

SplitAdapter factorizes the history representation into two context branches.
The object/load branch $f_{\mathrm{obj}}$, implemented as the Task Head in Fig.~\ref{fig:ocean_overview}, maps the shared history feature to an object/load latent, a mass estimate, and a loaded-state estimate, while the dynamics branch $f_{\mathrm{dyn}}$ maps the same feature to a dynamics-aware latent:
\begin{equation}
    (z_{\mathrm{obj},t}, \hat{m}_t, \hat{\ell}_t)
    =
    f_{\mathrm{obj}}(e_t),
    \qquad
    z_{\mathrm{dyn},t}
    =
    f_{\mathrm{dyn}}(e_t).
\end{equation}


Here, $z_{\mathrm{obj},t}$ and $z_{\mathrm{dyn},t}$ denote the object/load and dynamics-aware latents, and the effective payload estimate is defined as $m_{\mathrm{eff},t} = \hat{\ell}_t \hat{m}_t$.

\paragraph{Object/load context encoder.}
The object/load branch produces $z_{\mathrm{obj},t}$ while explicitly estimating object mass and loaded state. Its supervised loss is
\begin{equation}
    \mathcal{L}_{\mathrm{sup}}^{\mathrm{obj}}
    =
    \lambda_m \mathcal{L}_{\mathrm{mass}}
    +
    \lambda_{\ell} \mathcal{L}_{\mathrm{loaded}} .
\end{equation}
An optional sparse regularizer is applied as

\begin{equation}
    \mathcal{L}_{\mathrm{sparse}}^{\mathrm{obj}}
    =
    \lambda_{1}^{\mathrm{obj}}\|z_{\mathrm{obj},t}\|_1 .
\end{equation}

\paragraph{Dynamics-aware context encoder.}
The dynamics-aware branch produces $z_{\mathrm{dyn},t}$. We also apply an $L_1$ regularizer:

\begin{equation}
    \mathcal{L}_{\mathrm{sparse}}^{\mathrm{dyn}}
    =
    \lambda_{1}^{\mathrm{dyn}}\|z_{\mathrm{dyn},t}\|_1 .
\end{equation}

\subsection{Split World Models}
SplitAdapter uses world-model prediction as a proxy task for learning informative history embeddings, following the general idea of dynamics-aware adaptation in AnyAdapter~\cite{any2track}. Instead of using a single transition predictor, we split the prediction target into object transition and robot transition.

\paragraph{Object/load world model.}
The object/load world model predicts the object delta transition:
\begin{equation}
    {\Delta \hat{o}}^{\mathrm{obj}}_{t+1}
    =
    W_{\mathrm{obj}}
    (o_t, a_t, \hat{m}_t, \hat{\ell}_t, z_{\mathrm{obj},t}),
\end{equation}
where $o_t$ denotes the robot state used for transition prediction, and $\Delta p^{\mathrm{obj}}_t$ and $\Delta R^{\mathrm{obj}}_t$ denote the object position and orientation changes:
\begin{equation}
    \Delta p^{\mathrm{obj}}_t
    =
    p^{\mathrm{obj}}_{t+1} - p^{\mathrm{obj}}_t,
    \qquad
    \Delta R^{\mathrm{obj}}_t
    =
    R^{\mathrm{obj}}_{t+1}(R^{\mathrm{obj}}_t)^{-1}.
\end{equation}
The object/load world-model loss is
\begin{equation}
    \mathcal{L}_{\mathrm{objWM}}
    =
    \lambda_p
    \|\widehat{\Delta p}^{\mathrm{obj}}_t - \Delta p^{\mathrm{obj}}_t\|_1
    +
    \lambda_R
    d_R(\widehat{\Delta R}^{\mathrm{obj}}_t, \Delta R^{\mathrm{obj}}_t).
\end{equation}

\paragraph{Dynamics-aware world model.}
The dynamics-aware world model predicts the next robot state:
\begin{equation}
    \hat{o}_{t+1}
    =
    W_{\mathrm{dyn}}
    (o_t, a_t, z_{\mathrm{dyn},t}, m_{t}).
\end{equation}
The corresponding loss is
\begin{equation}
    \mathcal{L}_{\mathrm{dynWM}}
    =
    \|\hat{s}_{t+1} - s_{t+1}\|_1 .
\end{equation}
Both world models are conditioned on the estimated payload $m_{\mathrm{est},t}$, which reduces the need for the latent variables to redundantly encode payload mass. As a result, $z_{\mathrm{obj},t}$ is encouraged to emphasize object/load-related interaction context, while $z_{\mathrm{dyn},t}$ is encouraged to emphasize residual dynamics/environment mismatch.

\subsection{GRL-Based Cross-Adversarial Separation}
Although split world-model objectives encourage specialization, both context encoders share the same history feature $e_t$. To further reduce information leakage, we introduce a GRL-based cross-adversarial objective.
We use $A_{\mathrm{obj}\leftarrow\mathrm{dyn}}$ and $A_{\mathrm{robot}\leftarrow\mathrm{obj}}$ to denote adversarial predictors from the dynamics-aware latent to the object target and from the object/load latent to the robot target, respectively.

The first adversary predicts object transition from the dynamics-aware latent:
\begin{equation}
    {\Delta \hat{o}}^{\mathrm{obj,adv}}_{t+1}
    =
    A_{\mathrm{obj}\leftarrow\mathrm{dyn}}
    (s_t, a_t, \mathrm{GRL}(z_{\mathrm{dyn},t}), m_{\mathrm{est},t}).
\end{equation}
Through the gradient reversal layer, $z_{\mathrm{dyn},t}$ is discouraged from encoding object/load transition information. The second adversary predicts robot transition from the object/load latent:
\begin{equation}
    \hat{s}^{\mathrm{adv}}_{t+1}
    =
    A_{\mathrm{robot}\leftarrow\mathrm{obj}}
    (s_t, a_t, \mathrm{GRL}(z_{\mathrm{obj},t}), m_{\mathrm{est},t}).
\end{equation}
This discourages $z_{\mathrm{obj},t}$ from encoding robot dynamics and environment mismatch, and together with the split world models promotes branch specialization.

\subsection{Hierarchical FiLM-Based Feature Modulation}
SplitAdapter adapts the frozen base policy by modulating intermediate feature representations rather than directly perturbing the final action.
For a hidden activation $h$, Feature-wise Linear Modulation (FiLM)~\cite{perez2018film} applies
\begin{equation}
    h' = \gamma(z) \odot h + \beta(z),
\end{equation}
where $\gamma(z)$ and $\beta(z)$ are feature-wise scale and bias terms generated from context latents.
We apply hierarchical FiLM modulation across different layers of the frozen policy.
As shown in the lower part of Fig.~\ref{fig:ocean_overview}, object/load FiLM, generated from $(z_{\mathrm{obj},t}, m_{\mathrm{est},t})$, modulates earlier layers to shape coarse task-level behavior, such as payload-conditioned lifting posture and whole-body coordination.
In contrast, dynamics-aware FiLM, generated from $z_{\mathrm{dyn},t}$, modulates later layers closer to the action head to compensate for dynamics and environment mismatch.
This layer-dependent modulation allows the two adaptation factors to influence different levels of the frozen policy representation. The modulation layers are initialized to preserve the original frozen-policy behavior at the beginning of training, and adapter training uses the same reinforcement-learning objective and reward as the base policy.

\section{Experimental Results}
We design experiments to answer three questions. 
\textbf{Q1}: Does factorized adaptation improve humanoid loco-manipulation performance under payload and pickup/placement-height variations?
\textbf{Q2}: Does factorizing adaptation into object/load and dynamics-aware contexts outperform unified latent adaptation?
\textbf{Q3}: Can the proposed method improve sim-to-real transfer under heavy-load conditions?

\subsection{Simulation Evaluation}
We use a frozen PhysHSI-style AMP-based box manipulation policy as the base policy.
Training is performed in Isaac Gym~\cite{isaacgym}, and all results are verified through MuJoCo-based sim-to-sim transfer.

We compare the following methods:
\begin{itemize}[leftmargin=1.2em]
    \item \textbf{Base (PhysHSI~\cite{phys_hsi})}: frozen base policy.
    \item \mbox{\textbf{AnyAdapter-style WM-FiLM}:} world-model FiLM adapter following the AnyAdapter formulation~\cite{any2track}
    \item \textbf{Ours w/o split latent}: proposed model without latent splitting.
    \item \textbf{Ours w/o hierarchical FiLM}: proposed model without hierarchical FiLM.
    \item \textbf{Ours w/o GRL}: proposed model without GRL.
    \item \textbf{SplitAdapter}: full proposed method.
\end{itemize}

We evaluate three object masses (2, 4, and 6 kg) and three pickup/placement heights (0, 30, and 60 cm). Each condition is evaluated over ten trials.
We report \textbf{Lift-up success} and \textbf{Full-task success} counts.
Here, \textbf{Full-task success} denotes successful completion of the entire sequence of approach, lift, transport, and placement.

\subsubsection*{Performance Evaluation}

\newcommand{\score}[2]{\makebox[1.1em][r]{#1}/10\;\makebox[1.1em][r]{#2}/10}
\newcommand{\bscore}[2]{\makebox[1.1em][r]{\textbf{#1}}/10\;\makebox[1.1em][r]{\textbf{#2}}/10}
\newcommand{\bfirst}[2]{\makebox[1.1em][r]{\textbf{#1}}/10\;\makebox[1.1em][r]{#2}/10}
\newcommand{\bsecond}[2]{\makebox[1.1em][r]{#1}/10\;\makebox[1.1em][r]{\textbf{#2}}/10}



\begin{table*}[h]
\centering
\caption{
MuJoCo sim-to-sim Full-task success counts under object mass and pickup/placement-height variations.
Each condition is evaluated over ten trials.
Aggregate Full-task success and Lift-up success are summarized in the two rightmost columns.
Cell colors indicate higher-ranked results within each column.
The 6 kg, 0 cm setting is the most challenging case. 
}
\label{tab:simulation_combined}
\small
\setlength{\tabcolsep}{3.8pt}
\renewcommand{\arraystretch}{1.05}

\begin{adjustbox}{width=0.98\linewidth,center}
\begin{tabular}{@{}lccccccccc cc@{}}
\toprule
\multirow{2}{*}{Method}
& \multicolumn{3}{c}{2 kg}
& \multicolumn{3}{c}{4 kg}
& \multicolumn{3}{c}{6 kg}
& \multirow{2}{*}{Full-task}
& \multirow{2}{*}{Lift-up} \\
\cmidrule(lr){2-4}
\cmidrule(lr){5-7}
\cmidrule(lr){8-10}
& 0 cm & 30 cm & 60 cm
& 0 cm & 30 cm & 60 cm
& 0 cm & 30 cm & 60 cm
& & \\
\midrule

\multicolumn{12}{@{}l}{\textit{Baselines}} \\
Base (PhysHSI~\cite{phys_hsi})
& \first{10} & \third{8} & \second{9}
& \second{9} & \first{10} & \first{10}
& 5 & 5 & 5
& 71/90 & \third{86/90} \\

AnyAdapter-style WM-FiLM
& \first{10} & \first{10} & \second{9}
& \second{9} & \second{9} & \first{10}
& 4 & \third{6} & \second{8}
& 75/90 & \third{86/90} \\

\midrule
\multicolumn{12}{@{}l}{\textit{SplitAdapter variants}} \\

\rowcolor{gray!6}
\quad w/o split latent
& \first{10} & \second{9} & \first{10}
& \second{9} & \first{10} & \second{9}
& 6 & \first{9} & \first{9}
& \third{81/90} & \second{88/90} \\

\rowcolor{gray!6}
\quad w/o hierarchical FiLM
& \first{10} & \first{10} & \first{10}
& \second{9} & \first{10} & \second{9}
& \third{7} & \second{8} & \second{8}
& \third{81/90} & \first{90/90} \\

\rowcolor{gray!6}
\quad w/o GRL
& \first{10} & \third{8} & \first{10}
& \first{10} & \first{10} & \first{10}
& \second{8} & \first{9} & \first{9}
& \second{84/90} & \first{90/90} \\

\rowcolor{gray!12}
\textbf{SplitAdapter}
& \first{10} & \first{10} & \first{10}
& \second{9} & \first{10} & \first{10}
& \first{10} & \first{9} & \first{9}
& \first{86/90} & \first{90/90} \\

\bottomrule
\end{tabular}
\end{adjustbox}

\end{table*}

Table ~\ref{tab:simulation_combined} shows that SplitAdapter improves overall task performance, with higher success rates not only across the in-distribution mass settings but also on the 6 kg condition beyond the training range.
The advantage becomes clearer in the heavy load 6 kg regime, where stable transport and placement become substantially more difficult after lifting.
In particular, although Lift-up success often remains high, the base policy and the AnyAdapter-style WM-FiLM baseline frequently lose postural stability during transport and fail before completing placement.
In contrast, SplitAdapter maintains high success rates across all 6 kg pickup/placement-height variations, including the most challenging 6 kg, 0 cm condition.

\begin{figure}[h]
    \centering
    \begin{subfigure}{0.58\linewidth}
        \centering
        \includegraphics[width=\linewidth]{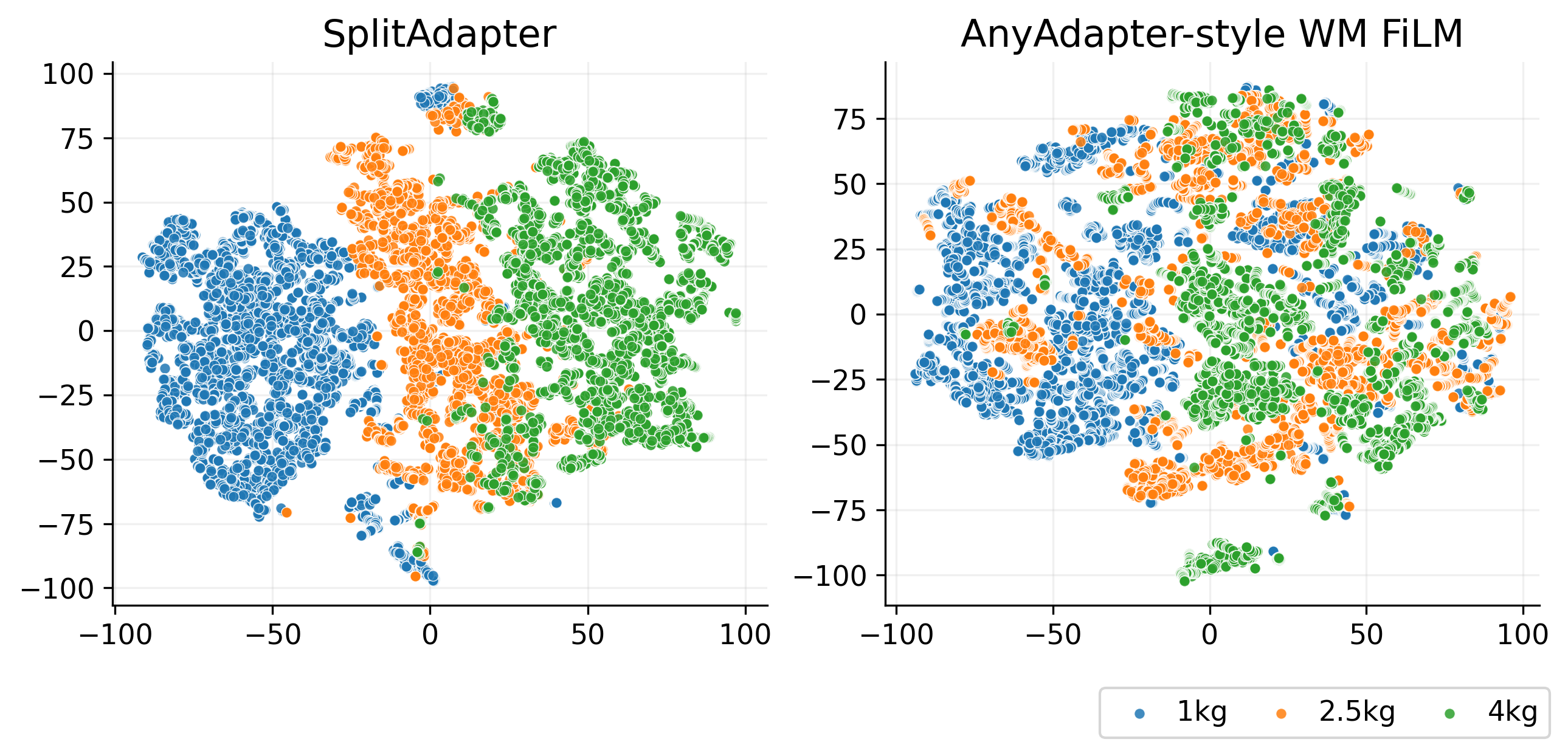}
        \caption{}
        \label{fig:fig8}
    \end{subfigure}\hfill
    \begin{subfigure}{0.30\linewidth}
        \centering
        \includegraphics[width=\linewidth]{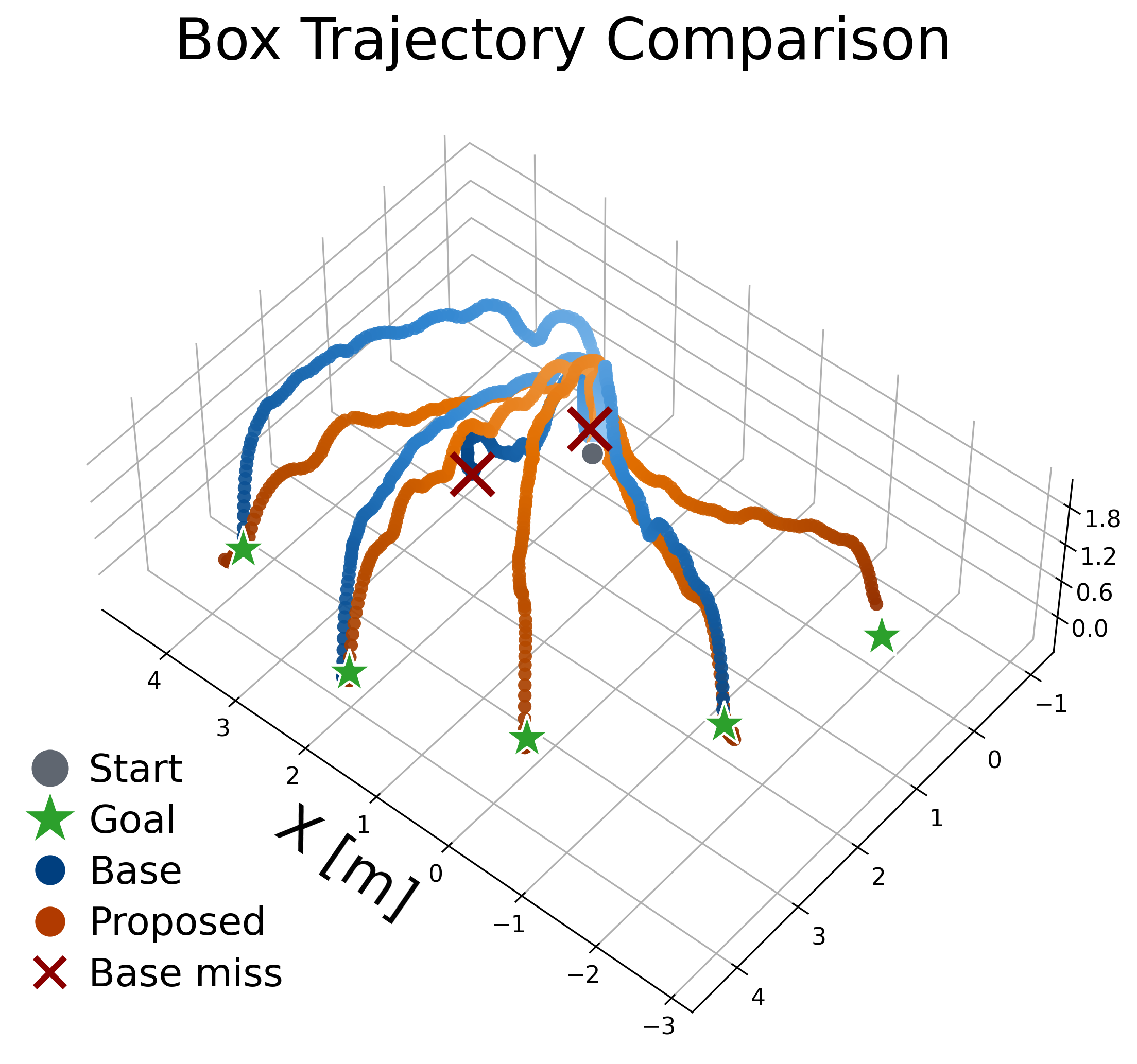}
        \caption{}
        \label{fig:fig_traj_comp}
    \end{subfigure}
    \caption{Qualitative comparisons of the proposed and baseline methods. (a) t-SNE visualization of adaptation representations across object masses for the proposed split latent $[z_{\mathrm{obj}}; z_{\mathrm{dyn}}]$ and the unified latent baseline. (b) Box-transport trajectories for a 6 kg, 0 cm floor-lift task under multiple goal directions. The base policy (PhysHSI~\cite{phys_hsi}) exhibits unstable transport, including failure cases and less direct paths to the target due to load-induced balance disturbances.}
    \label{fig:fig_tsne_and_traj}
\end{figure}

\begin{figure}[h]
    \centering
    \includegraphics[width=0.9\linewidth]{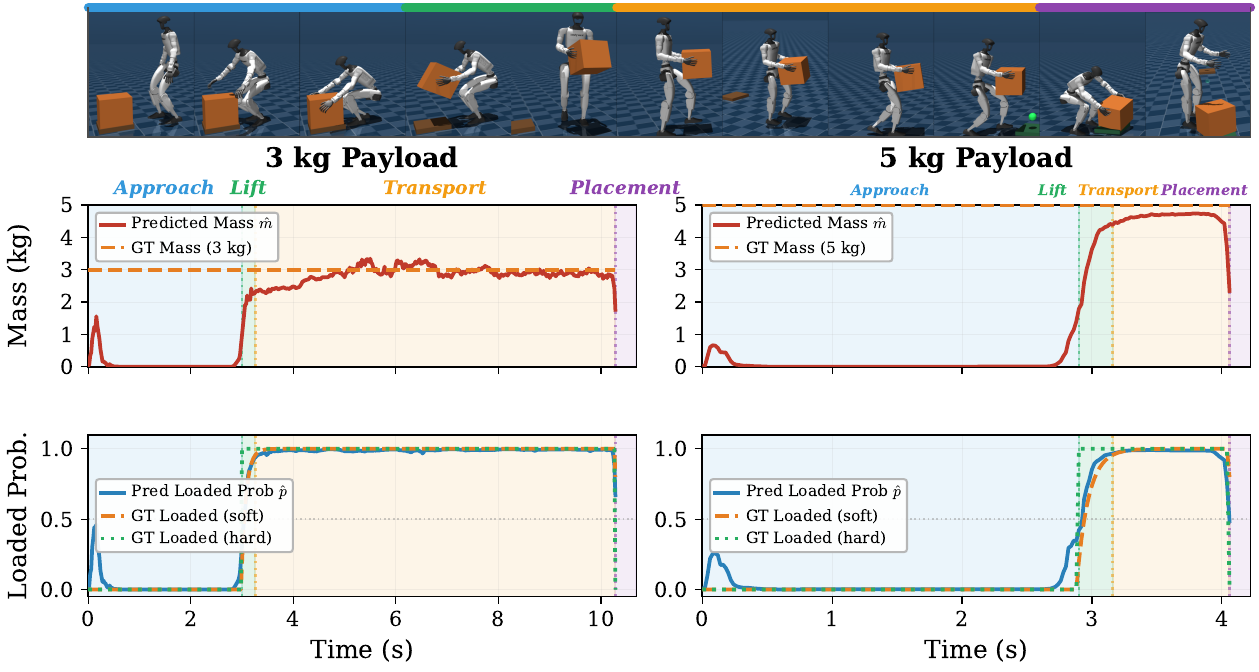}
    \caption{Online mass and loaded-state estimation across two payload conditions. Both the estimated mass and loaded probability quickly converge to the ground truth after lifting. Soft and hard labels denote probabilistic and binary loaded-state targets, respectively.}
    \label{fig:fig_mass_est}
\end{figure}

The performance gap between SplitAdapter and the baselines becomes more pronounced as object mass increases, indicating that the proposed adaptation is particularly effective under stronger load-induced dynamics changes. Compared with the AnyAdapter-style WM-FiLM baseline,
which uses a unified latent, SplitAdapter exhibits clearer mass-dependent organization in the latent space, as shown in Fig.~\ref{fig:fig_tsne_and_traj} (a). This suggests that the factorized branch structure preserves load-related variation more distinctly. In addition, SplitAdapter directly uses the estimated mass
and loaded state as adaptation signals, as shown in Fig.~\ref{fig:fig_mass_est}, which further helps the policy respond to changes in object weight. As illustrated in Fig.~\ref{fig:fig_tsne_and_traj} (b), the proposed method generalizes across diverse goal directions, while the base policy often deviates from direct paths or fails due to load-induced disturbances.

\subsubsection*{Effect of Latent Splitting and GRL}



\begin{figure}[h]
    \centering
    \includegraphics[width=0.8\linewidth]{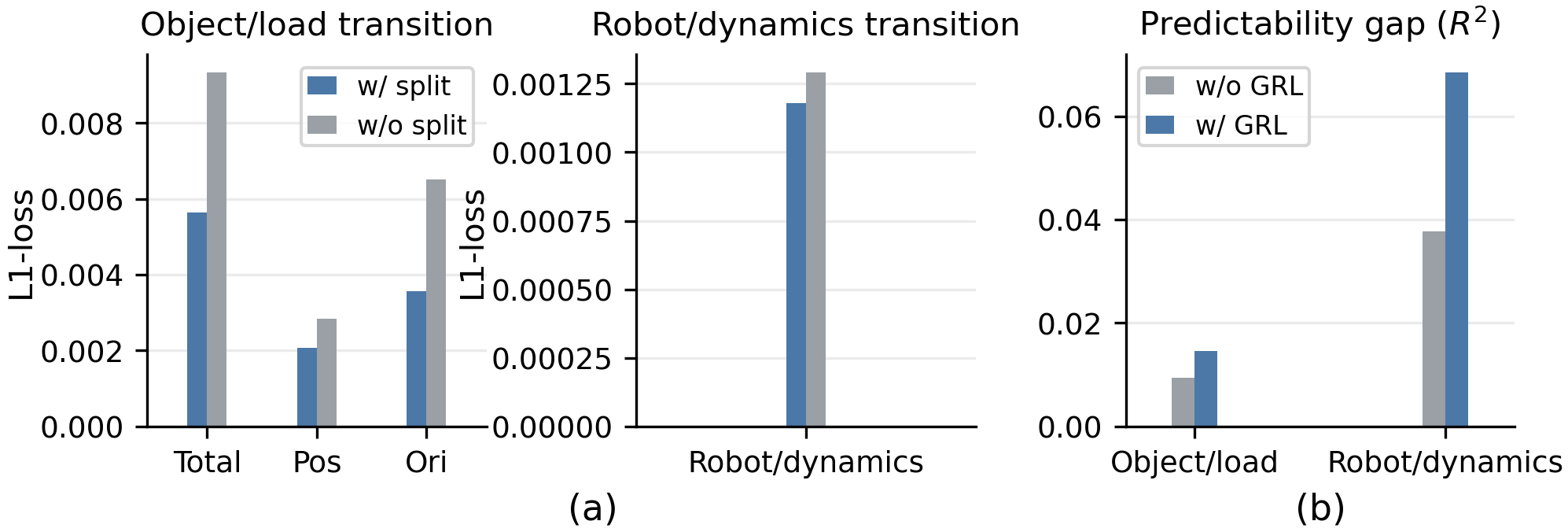}
    \caption{(a) World-model transition losses with and without split designs. (b) Predictability gaps with and without GRL for object/load and robot/dynamics transitions.}
    \label{fig:fig5_combined}
\end{figure}

Among the SplitAdapter variants, removing latent splitting yields the lowest Full-task performance, especially in the 6 kg regime.
This suggests that, even when mass and loaded-state information are provided, a unified latent tends to mix load-related information and dynamics-related signals, making it harder for the adapter to emphasize their different predictive roles.
This interpretation is supported by Fig.~\ref{fig:fig5_combined} (a), where the factorized design yields lower world-model transition losses for both the object/load and robot-transition targets.



Removing GRL causes a smaller but still consistent drop compared with the full model, suggesting that cross-adversarial regularization helps the two branches remain sufficiently specialized to better exploit the factorized adaptation structure. As shown in
Fig.~\ref{fig:fig5_combined} (b), the predictability gap between matched-target and cross-target prediction becomes larger when GRL is used. Here, the predictability gap is defined as the difference in probe regression $R^2$ between the matched branch-target pair and the cross-branch pair for each transition target. Larger gaps indicate that the intended branch explains its assigned transition more strongly than the opposite
branch.

Removing hierarchical FiLM also reduces Full-task success relative to the full model, while preserving high Lift-up success.
This pattern suggests that hierarchical FiLM is not critical for initiating the lift itself, but is beneficial for subsequent whole-body coordination during transport and placement under changing load conditions.

\subsection{Real-World Evaluation}

We evaluate the Base policy and SplitAdapter on the Unitree G1~\cite{unitree_g1} under zero-shot real-world deployment. Each method is tested on 9 conditions with three trials per condition, without additional fine-tuning.

\begin{table}[!h]
\vspace{-4mm}
\centering
\caption{
Real-world deployment results under object mass and pickup/placement-height variations. Each entry reports Full-task success over three trials, with Lift-up success shown in parentheses.}
\label{tab:real_grid}
\small
\setlength{\tabcolsep}{3.6pt}
\renewcommand{\arraystretch}{1.05}

\begin{adjustbox}{width=0.98\linewidth,center}
\begin{tabular}{@{}lccccccccc cc@{}}
\toprule
\multirow{2}{*}{Method}
& \multicolumn{3}{c}{2 kg}
& \multicolumn{3}{c}{4 kg}
& \multicolumn{3}{c}{6 kg}
& \multirow{2}{*}{Full-task}
& \multirow{2}{*}{Lift-up} \\
\cmidrule(lr){2-4}
\cmidrule(lr){5-7}
\cmidrule(lr){8-10}
& 0 cm & 30 cm & 60 cm
& 0 cm & 30 cm & 60 cm
& 0 cm & 30 cm & 60 cm
& & \\
\midrule

Base (PhysHSI~\cite{phys_hsi})
& 3 (3) & 3 (3) & 3 (3)
& 1 (2) & 2 (3) & 2 (2)
& 1 (2) & 1 (3) & 0 (1)
& 16/27 (59.3\%) & 22/27 (81.5\%) \\

\rowcolor{gray!12}
\textbf{SplitAdapter}
& \textbf{3 (3)} & \textbf{3 (3)} & \textbf{3 (3)}
& \textbf{3 (3)} & \textbf{3 (3)} & \textbf{3 (3)}
& \textbf{2 (3)} & \textbf{3 (3)} & \textbf{3 (3)}
& \textbf{26/27 (96.3\%)} & \textbf{27/27 (100.0\%)} \\

\bottomrule
\end{tabular}
\end{adjustbox}

\vspace{-0.5em}
\end{table}

Table~\ref{tab:real_grid} shows that the performance improvement observed in simulation transfers to the real robot under zero-shot deployment.
The Base policy succeeds reliably in the lighter 2 kg conditions, but its performance deteriorates as object mass increases, yielding 16/27 Full-task successes and 22/27 Lift-up successes overall.
In particular, the failure gap becomes more pronounced in the 6 kg settings, where the policy can still initiate lifting in several trials but frequently loses grasp stability, torso alignment, or whole-body balance before completing transport and placement.
In contrast, SplitAdapter achieves 26/27 Full-task successes and 27/27 Lift-up successes, while remaining robust across all 4 kg conditions and nearly all 6 kg conditions.
These results indicate that the proposed factorized adaptation improves not only Lift-up success but, more importantly, the stability of the subsequent transport and placement phases in real-world heavy-load deployment.

\section{Conclusion}

We presented \textbf{SplitAdapter: Load-Aware Humanoid Loco-Manipulation via Factorized Adaptation}, a factorized adaptation framework for heavy-load humanoid box manipulation. 
SplitAdapter augments a frozen manipulation policy with object/load-aware and dynamics-aware context encoders, split world-model objectives, GRL-based regularization, and hierarchical FiLM modulation. 
Experimental results show improved robustness under mass and pickup/placement-height variations in both simulation and real-world deployment, particularly in the heaviest 6 kg setting.

\section{Limitations}

While the proposed framework shows strong performance in simulation and real-world evaluation, several limitations remain.
The current study focuses on rigid box lifting, transport, and placement, and does not address deformable objects or broader contact-rich manipulation scenarios.
In addition, the scale of the real-world evaluation remains limited due to hardware cost and durability constraints under repeated heavy-load testing, since the Unitree G1~\cite{unitree_g1} arm payload is specified at approximately 3 kg per arm.
\FloatBarrier
\clearpage
\bibliography{references}

\clearpage

\appendix
\section{Appendix}

\subsection{Observation and Reward Details}
\noindent The frozen base policy follows the AMP-based locomotion framework of PhysHSI~\cite{phys_hsi}. The policy outputs 29-D joint-position targets executed through low-level PD control. Here, we only summarize the task-specific observations and rewards used in our scenario. Entries marked with `*' denote additions beyond PhysHSI~\cite{phys_hsi}.

\paragraph{Task Observations.}
The task-specific observation $o_t^{G}\in\mathbb{R}^{15}$ comprises the following properties of the target box:
\begin{itemize}[leftmargin=1.2em, itemsep=0pt, topsep=2pt]
\item local box position $p_t^{o}\in\mathbb{R}^{3}$,
\item local box rotation $R_t^{o}\in\mathbb{R}^{6}$,
\item box size $b_t\in\mathbb{R}^{3}$,
\item local goal position $p_t^{g}\in\mathbb{R}^{3}$.
\end{itemize}
The one-step policy observation has 123 dimensions, consisting of 108 proprioceptive dimensions and the 15-D task observation. The privileged critic observation has 126 dimensions, obtained by augmenting the actor-side observation with the 3-D base linear velocity. SplitAdapter additionally consumes a 50-step adaptation history$^{*}$ of 123-D observations.

\paragraph{Task Rewards.}
Following PhysHSI~\cite{phys_hsi}, the task reward is decomposed into approach, lift, relocation, and stabilization stages. We report only the active task-specific terms used in our experiments. The total task reward is
\begin{equation}
    r_t^{G} = r_t^{\mathrm{walk}} + r_t^{\mathrm{carryup}} + r_t^{\mathrm{relocation}} + r_t^{\mathrm{hold}*} + r_t^{\mathrm{standup}*}.
\end{equation}
Relative to PhysHSI~\cite{phys_hsi}, the additional nonzero task terms are $r_t^{\mathrm{hold}*}$ and $r_t^{\mathrm{standup}*}$.
\begin{equation}
    r_t^{\mathrm{walk}} =
    \begin{cases}
        1.5, & d_{ro}<0.7,\\
        r_{ro,\mathrm{vel}} + r_{ro,\mathrm{heading}}, & \text{otherwise},
    \end{cases}
\end{equation}
\begin{equation}
    \begin{aligned}
    r_t^{\mathrm{carryup}} ={}& 0.7r_t^{\mathrm{hand\_pos}} + 0.7r_t^{\mathrm{hand\_contact}}\\
    &+ 2.0r_t^{\mathrm{box\_height}} + 0.3r_t^{\mathrm{pelvis\_box}},
    \end{aligned}
\end{equation}
\begin{equation}
    \begin{aligned}
    r_t^{\mathrm{relocation}} ={}& 1.0r_{rg,\mathrm{vel}} + 0.5r_t^{\mathrm{reloc\_heading}}\\
    &+ 1.0r_{og,\mathrm{pos}} + 1.0r_t^{\mathrm{put}},
    \end{aligned}
\end{equation}
\noindent Here, $d_{ro}$, $d_{og}$, and $v_{ro}, v_{rg}$ denote planar distances and projected velocities, and $\psi_{ro}^{\star}, \psi_{rg}^{\star}$ denote desired yaw directions. As in PhysHSI~\cite{phys_hsi}, $r_{ro,\mathrm{vel}}=\exp(-5(0.85-v_{ro})^2)$ and $r_{ro,\mathrm{heading}}$ is the average pelvis/torso heading-alignment term. Likewise, $r_t^{\mathrm{carryup}}$ combines hand-position, hand-contact, lift-height, and pelvis-alignment terms, while $r_t^{\mathrm{relocation}}$ combines commanded transport velocity, heading alignment, object-goal distance, and vertical placement.
\begin{equation}
    r_t^{\mathrm{hand\_pos}}=\exp\!\left(-3\|\mathbf{p}_{\mathrm{hand,avg}}-\mathbf{p}_{\mathrm{box}}\|_2^2\right), \quad
    r_t^{\mathrm{box\_height}}=\exp\!\left(-3\max(h_{\mathrm{box}}^\star-h_{\mathrm{box}},0)\right),
\end{equation}
\begin{equation}
    \begin{aligned}
    r_t^{\mathrm{hand\_contact}}
    ={}& \min\!\left(\frac{\min_k f_{\mathrm{hand},k}}{20},1\right)
    \mathbb{1}\!\left[\exists k,\;f_{\mathrm{wrist},k}>1.0\right],\\
    r_t^{\mathrm{pelvis\_box}}
    ={}& \exp(-0.75|\mathrm{wrap}(\psi_{ro}^{\star}-\psi_{\mathrm{pelvis}})|),
    \end{aligned}
\end{equation}
\begin{equation}
    r_{rg,\mathrm{vel}}=\exp(-5(v_{rg}-0.85)^2), \quad
    r_t^{\mathrm{reloc\_heading}}=\exp(-0.75|\mathrm{wrap}(\psi_{rg}^{\star}-\psi)|),
\end{equation}
\begin{equation}
    r_{og,\mathrm{pos}}=\exp(-10d_{og}), \quad
    r_t^{\mathrm{put}}=\exp(-3|z_{\mathrm{box}}-z_{\mathrm{goal}}|).
\end{equation}
Our implementation further adds a hold-phase stabilization reward and a post-success stabilization reward:
\begin{equation}
    r_t^{\mathrm{hold}*}
    =
    \exp(-0.3e_{\mathrm{stand}})
    -
    0.05\lVert \boldsymbol{\omega}_{\mathrm{pelvis},xy}\rVert_2^2
    -
    0.02\lVert \boldsymbol{\omega}_{\mathrm{torso},xy}\rVert_2^2,
\end{equation}
\begin{equation}
    \begin{aligned}
    r_t^{\mathrm{standup}*}
    ={}& 0.5r_t^{\mathrm{head\_height}}
    + r_t^{\mathrm{standup\_heading}}
    + 1.5\exp(-0.3e_{\mathrm{stand}})\\
    &+ 0.5r_t^{\mathrm{hand\_free}}
    - 0.05\lVert \boldsymbol{\omega}_{\mathrm{pelvis},xy}\rVert_2^2
    - 0.02\lVert \boldsymbol{\omega}_{\mathrm{torso},xy}\rVert_2^2.
    \end{aligned}
\end{equation}
Here, $r_t^{\mathrm{hand\_free}}$ rewards release after placement: it takes value $1$ when neither hand remains in contact with the box, $0.5$ when only one hand remains in contact, and $0$ when both hands are still in contact.
The only additional nonzero regularizers beyond PhysHSI~\cite{phys_hsi} are \texttt{ankle\_roll\_close}$^*$ and \texttt{low\_box\_time\_penalty}$^*$:
\begin{equation}
    r_{\mathrm{ankle\_roll\_close}} = \left[\max(d_{\mathrm{th}}^{\mathrm{ankle}} - d_{\mathrm{ankle},y}, 0)\right]^2,
\end{equation}
\begin{equation}
    T_{\mathrm{low},t} =
    \begin{cases}
        T_{\mathrm{low},t-1} + \Delta t, & \text{if the box is near and too low},\\
        0, & \text{otherwise},
    \end{cases}
\end{equation}
\begin{equation}
    r_{\mathrm{low\_box\_time\_penalty}} = \max(T_{\mathrm{low},t} - T_{\mathrm{grace}}, 0),
\end{equation}
where $T_{\mathrm{grace}}=2.0$ s. The first term penalizes small lateral ankle spacing, and the second penalizes prolonged failure cases in which the robot remains near the object while keeping the box too low.
\FloatBarrier

\subsection{Domain Randomization}
\paragraph{Task Setup Randomization.}
We randomize the carry-box task setup along the following dimensions:
\begin{itemize}[leftmargin=1.2em, itemsep=0pt, topsep=2pt]
\item the 2D positions of the box and target are sampled relative to the initial base position,
\item the pickup and placement heights are sampled within $[0.0, 0.6]$ m above the ground,
\item the box size is randomized with axis-wise ranges $x\in[0.25,0.45]$ m, $y\in[0.25,0.45]$ m, and $z\in[0.25,0.35]$ m,
\item the box mass is randomized within the training range of $[0.5, 5.0]$ kg.
\end{itemize}

\begin{table}[ht]
\centering
\caption{Parameters of domain randomization.}
\label{tab:app_dr}
\scriptsize
\setlength{\tabcolsep}{3pt}
\renewcommand{\arraystretch}{1.04}
\begin{adjustbox}{width=0.9\linewidth,center}
\begin{tabular}{@{}p{0.20\linewidth}p{0.26\linewidth}p{0.40\linewidth}@{}}
\toprule
Category & Term & Value \\
\midrule
\multirow[t]{6}{*}{\textit{Physical Properties}} & actuator offset & $U(-0.05, 0.05)$ rad \\
 & motor strength & $U(0.9, 1.1)$ \\
 & center of mass & $U(-0.1, 0.1)$ m \\
 & link mass & $U(0.8, 1.2)$ \\
 & $K_p$, $K_d$ & $U(0.9, 1.1)$ \\
 & external disturbance & \makecell[l]{every 8 steps,\\magnitude in $[-50, 50]$} \\
\midrule
\multirow[t]{4}{*}{\textit{Observation Noise}} & angular velocity & 0.3 \\
 & projected gravity & 0.05 \\
 & joint position & 0.02 \\
 & joint velocity & 2.0 \\
\midrule
\multirow[t]{2}{*}{\textit{Task Observation Noise}} & box/goal position & $\pm 0.05$ m \\
 & box rotation & $\pm 5^\circ$ \\
\midrule
\multirow[t]{3}{*}{\textit{Object Dynamics}} & box friction & $U(0.5, 1.0)$ \\
 & box restitution & $U(0.05, 0.2)$ \\
 & platform friction & $U(0.5, 2.0)$ \\
\bottomrule
\end{tabular}
\end{adjustbox}
\end{table}

\clearpage
\subsection{Training Hyperparameters and Network Architecture}
\begin{table}[h]
\centering
\caption{Hyperparameters for training.}
\label{tab:app_train_hparams}
\scriptsize
\setlength{\tabcolsep}{3pt}
\renewcommand{\arraystretch}{1.05}
\begin{adjustbox}{width=0.9\linewidth,center}
\begin{tabular}{@{}p{0.22\linewidth}p{0.28\linewidth}p{0.32\linewidth}@{}}
\toprule
Category & Hyperparameter & Value \\
\midrule
\multirow[t]{8}{*}{\textit{General}} & num of envs & 4096 \\
 & num of steps per iteration & 100 \\
 & num of epochs & 5 \\
 & clip range & 0.2 \\
 & discount factor $\gamma$ & 0.99 \\
 & GAE balancing factor $\lambda$ & 0.95 \\
 & desired KL-divergence & 0.003 \\
 & actor and critic NN & MLP, hidden units $[512, 256, 256]$ \\
\midrule
\multirow[t]{6}{*}{\textit{Adaptation module}} & adapter history length & 50 \\
 & auxiliary learning rate & $1\times10^{-3}$ \\
 & FiLM hidden dimension & 128 \\
 & object / dynamics latent size & 16 / 16 \\
 & GRL coefficient / weight & 0.1 / $1\times10^{-3}$ \\
 & GRL warmup steps & 10000 \\
\bottomrule
\end{tabular}
\end{adjustbox}
\end{table}

\begin{table}[h]
\centering
\caption{Network architecture of the base policy and adaptation modules.}
\label{tab:app_adapter_arch}
\scriptsize
\setlength{\tabcolsep}{2pt}
\renewcommand{\arraystretch}{1.04}
\begin{adjustbox}{width=0.9\linewidth,center}
\begin{tabular}{@{}p{0.18\linewidth}p{0.22\linewidth}p{0.40\linewidth}@{}}
\toprule
Module & Component & Configuration \\
\midrule
\multirow[t]{1}{*}{Base policy} & Actor / critic & \makecell[l]{MLP $[512, 256, 256]$ /\\MLP $[512, 256, 256]$} \\
\midrule
\multirow[t]{6}{*}{Adaptation module} & Adapter history & 50 steps of 123-D observations \\
 & Shared temporal encoder & \makecell[l]{Conv1d$(123,64,k=6,s=2)$ + ELU\\Conv1d$(64,32,k=4,s=2)$ + ELU} \\
 & Shared feature dimension & 320 \\
 & Object/load branch & \makecell[l]{Linear$(320,64)$ + ELU\\Linear$(64,64)$ + ELU} \\
 & Dynamics branch & \makecell[l]{Linear$(320,64)$ + ELU\\Linear$(64,16)$} \\
 & FiLM generator & \makecell[l]{hidden size 128,\\soft-split routing} \\
\midrule
\multirow[t]{4}{*}{Latents / predictors} & Object latent & $z_{\mathrm{obj}}\in\mathbb{R}^{16}$ \\
 & Dynamics latent & $z_{\mathrm{dyn}}\in\mathbb{R}^{16}$ \\
 & Object world model & \makecell[l]{MLP $[64, 64]$,\\object delta target} \\
 & Robot world model & \makecell[l]{MLP $[128, 128]$,\\79-D robot transition target} \\
\bottomrule
\end{tabular}
\end{adjustbox}
\end{table}

\begin{table}[ht]
\centering
\caption{Training losses and regularizers for the adaptation modules.}
\label{tab:app_aux_objectives}
\scriptsize
\setlength{\tabcolsep}{3pt}
\renewcommand{\arraystretch}{1.04}
\begin{adjustbox}{width=0.9\linewidth,center}
\begin{tabular}{@{}p{0.42\linewidth}p{0.18\linewidth}p{0.28\linewidth}@{}}
\toprule
Objective & Weight & Description \\
\midrule
Mass regression $\mathcal{L}_{\mathrm{mass}}$ & 2.0 & box-mass estimation \\
Loaded-state BCE $\mathcal{L}_{\mathrm{loaded}}$ & 1.0 & loaded-state prediction \\
Object world model $\mathcal{L}_{\mathrm{objWM}}$ & 1.0 & object delta prediction \\
Robot world model $\mathcal{L}_{\mathrm{dynWM}}$ & 1.0 & robot transition prediction \\
\midrule
GRL adversarial loss $\mathcal{L}_{\mathrm{GRL}}$ & $1\times10^{-3}$ & cross-branch disentanglement \\
KL regularization $\mathcal{L}_{\mathrm{KL}}$ & $1\times10^{-3}$ & latent regularization \\
Object-latent $L_1$ norm $\|z_{\mathrm{obj}}\|_1$ & $1\times10^{-3}$ & sparsity on object latent \\
Dynamics-latent $L_1$ norm $\|z_{\mathrm{dyn}}\|_1$ & $5\times10^{-3}$ & sparsity on dynamics latent \\
\midrule
GRL coefficient & 0.1 & gradient-reversal scale \\
GRL warmup steps & 10000 & delayed adversarial activation \\
\bottomrule
\end{tabular}
\end{adjustbox}
\end{table}
\FloatBarrier

\clearpage
\subsection{Experiment Details}

\subsubsection*{Simulation Protocol}

Simulation evaluation is conducted under object masses of $2$, $4$, and $6$\,kg and pickup/placement heights of $0$, $30$, and $60$\,cm, as summarized in Table~\ref{tab:simulation_combined}. The success condition follows the same \textbf{Lift-up success} and \textbf{Full-task success} criteria as in the main text. Since training randomizes box mass only within $[0.5,5.0]$\,kg, the 6\,kg condition is treated as an out-of-distribution evaluation setting. During training, the success condition requires both a limited box tilt, measured by the norm of the box projected-gravity vector in the horizontal plane, $\|\mathbf{g}^{xy}_{\mathrm{box}}\|<0.1$, and a final object-to-goal distance below $0.1$\,m. 


\FloatBarrier

\subsubsection*{Real-World Protocol}

The real-world evaluation focuses on zero-shot deployment of the learned controller on the Unitree G1 platform~\cite{unitree_g1}. We evaluate the 9 condition combinations reported in Table~\ref{tab:real_grid}, using a validation box with fixed geometry $0.30 \times 0.43 \times 0.30$\,m to isolate the effects of object mass and pickup/placement-height variations.

In each trial, the robot approaches a box placed approximately 2\,m in front of its initial position, lifts the box, and transports it to a goal pose located 4\,m behind the pickup location.

\begin{figure}[h]
    \centering
    \includegraphics[width=0.7\columnwidth]{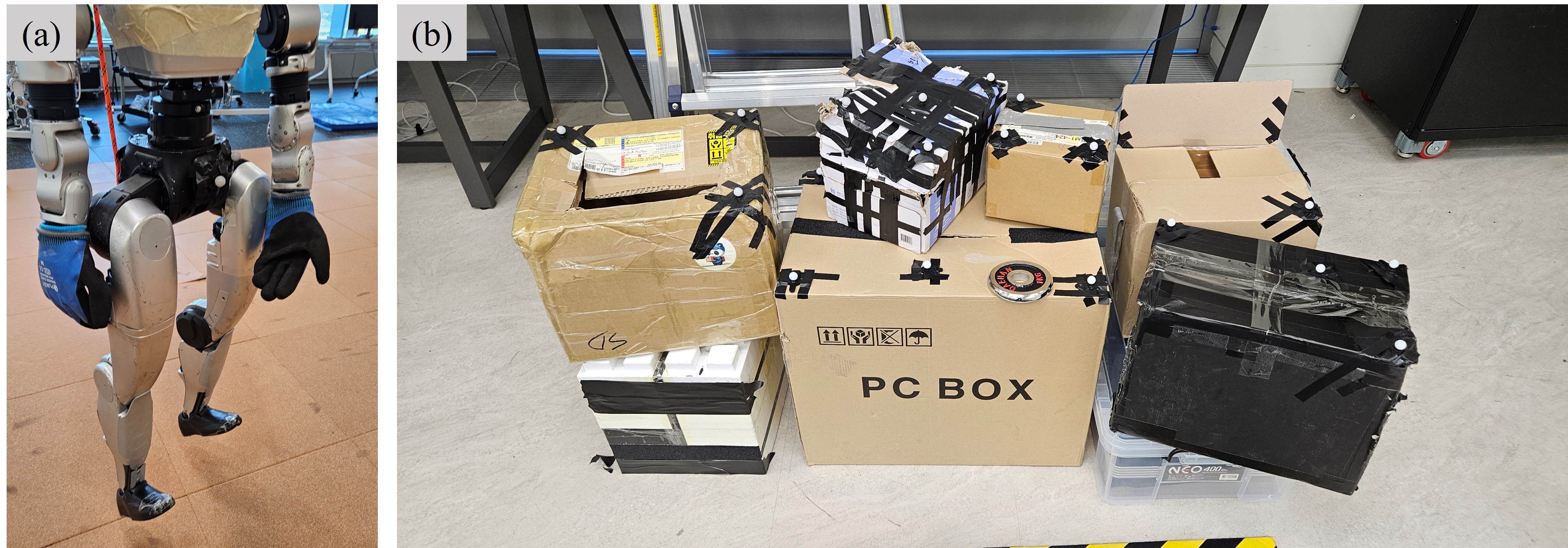}
    \caption{
    Real-world experiment setup.
    (a) High-friction gloves attached to the Unitree G1 hands to improve grasp contact reliability during box lifting and transport.
    (b) Validation boxes and payload configurations used for real-world evaluation. The box geometry is fixed while the payload mass and pickup/placement height are varied.
    }
    \label{fig:real_setup}
\end{figure}

To improve contact reliability during grasping and transport, the robot uses high-friction gloves, as shown in Fig.~\ref{fig:real_setup}(a). We use an external motion-capture system to obtain the robot and box poses required by the policy observation; motion-capture markers are attached to the robot pelvis and the top of the box, as illustrated in Fig.~\ref{fig:real_setup}. Additional qualitative evaluations are conducted using the box family shown in Fig.~\ref{fig:real_setup}(b), including the large-box and acrylic-box examples presented in Fig.~\ref{fig:firstpage}(c).

\clearpage

\subsection{Additional Simulation Results}

\begin{figure}[h]
    \centering
    \includegraphics[width=0.9\columnwidth]{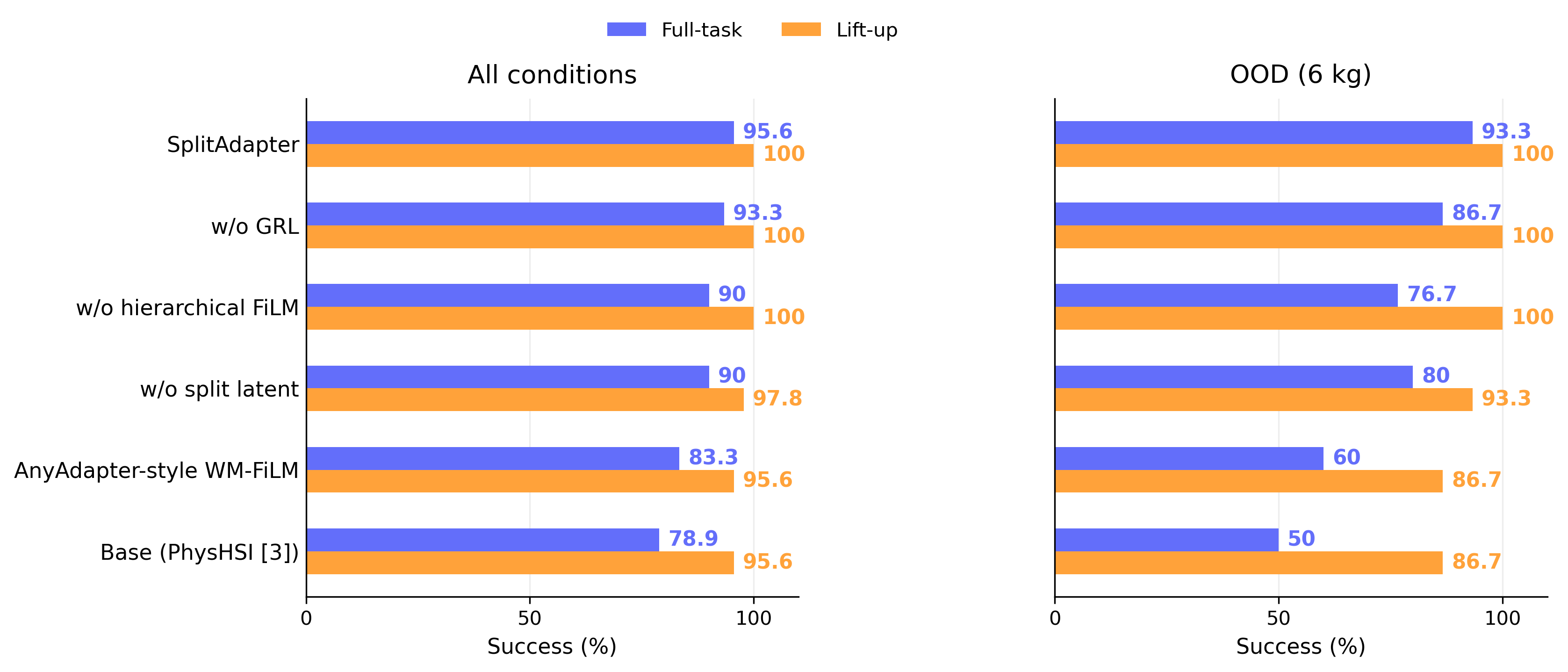}
    \caption{
    Supplementary visualization of the simulation results in Table~\ref{tab:simulation_combined}. The left plot summarizes overall Lift-up and Full-task success across all evaluation conditions, while the right plot focuses on the OOD 6\,kg setting.
    }
    \label{fig:fig10_table1_summary}
\end{figure}

\begin{figure}[h]
    \centering
    \includegraphics[width=0.8\columnwidth]{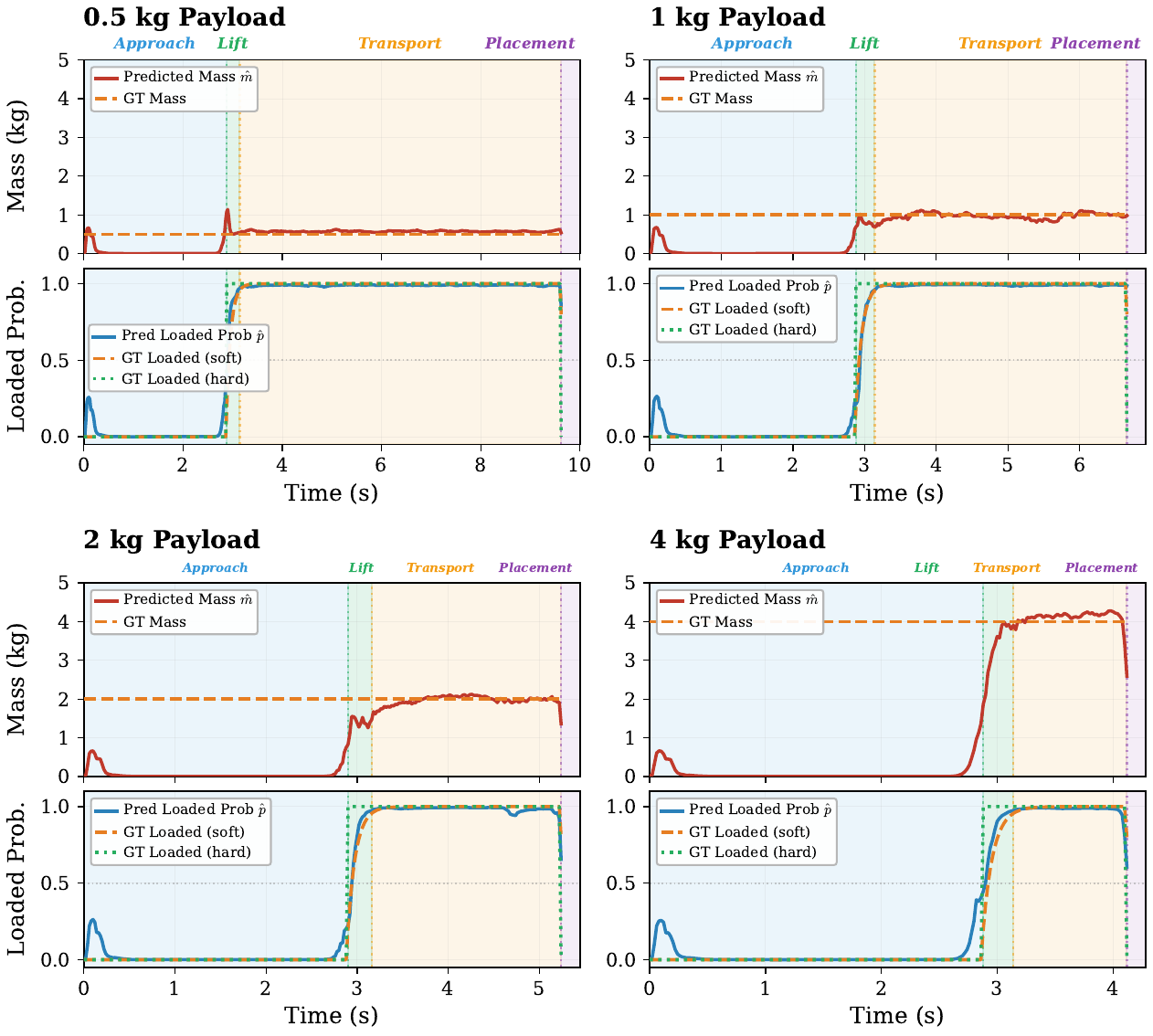}
    \caption{
    Additional online mass and loaded-state estimation results across four payload conditions (0.5\,kg, 1\,kg, 2\,kg, and 4\,kg). Consistent with the main results, the predicted mass $\hat{m}$, weighted by the loaded probability, converges toward the corresponding ground-truth payload after lifting and remains stable during transport. The loaded probability $\hat{p}$ reliably captures the transition from unloaded to loaded states and closely follows both the soft and hard ground-truth labels across all payload conditions.
    }
    \label{fig:fig_mass_est_full}
\end{figure}

\begin{figure}[h]
    \centering
    \includegraphics[width=\columnwidth]{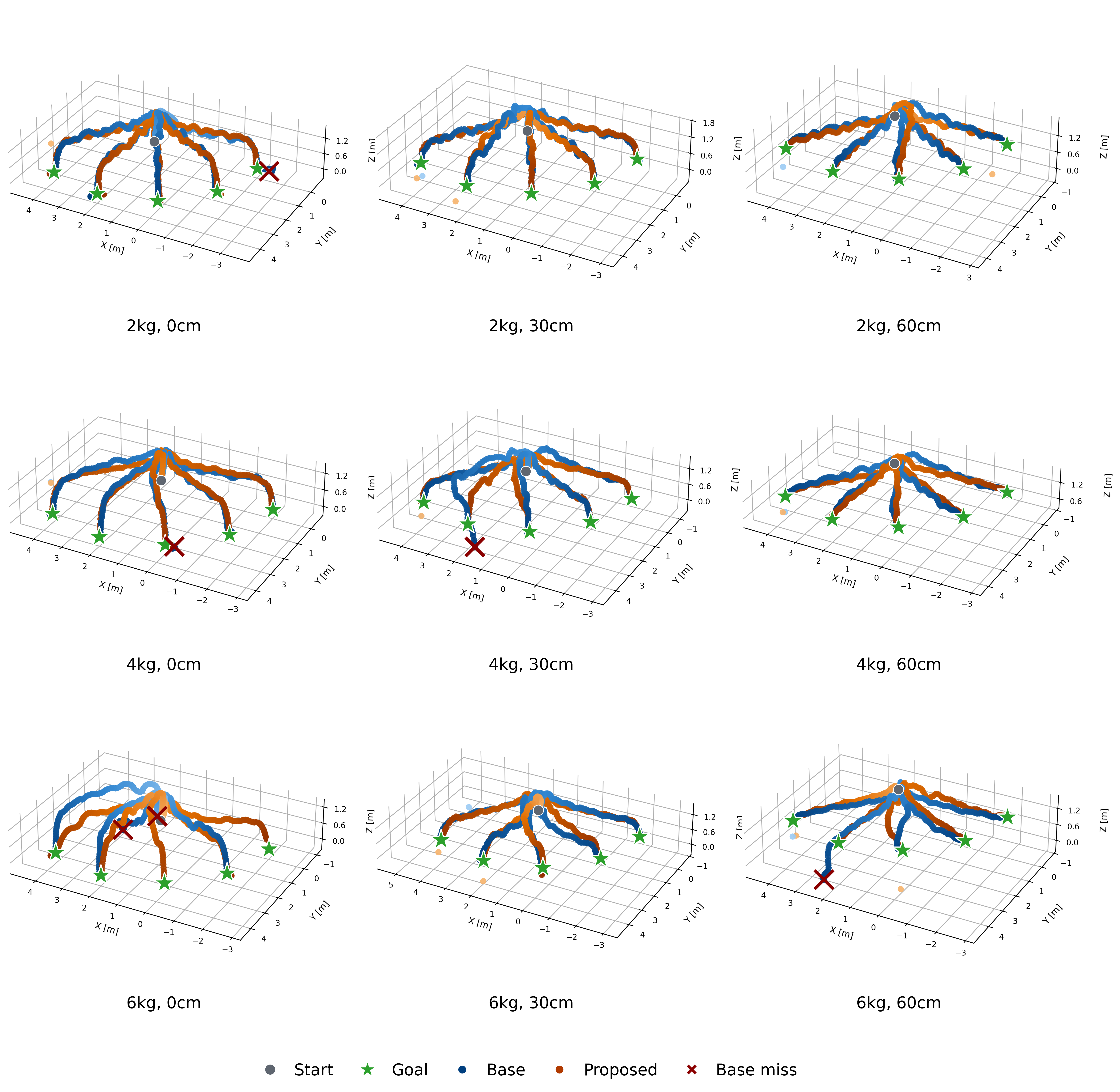}
    \caption{
Box-transport trajectories under 9 object mass and pickup/placement-height conditions, evaluated with multiple goal directions.
Blue and orange curves denote the base policy and SplitAdapter, respectively.
Green stars indicate target locations, and red crosses indicate failed trials of the base policy.
The proposed method maintains stable transport trajectories across diverse goals, whereas the base policy exhibits larger path deviations and more frequent failures under heavier payloads.    }
    \label{fig:fig_mass_est_full}
\end{figure}

\FloatBarrier



\end{document}